\documentclass[10pt,journal,compsoc]{IEEEtran}
\IEEEoverridecommandlockouts
\PassOptionsToPackage{prologue,dvipsnames}{xcolor}
\usepackage{algorithm}
\usepackage{amssymb}
\usepackage{amsmath,amsfonts}
\usepackage{array}
\usepackage{url}
\usepackage{subfig}
\usepackage{overpic}
\usepackage{threeparttable}
\usepackage{booktabs}
\usepackage{diagbox}
\usepackage{makecell}
\usepackage{colortbl}
\usepackage{color, soul}
\usepackage{multirow}
\usepackage{bbding}
\usepackage{xspace}
\usepackage{verbatim}
\usepackage{paralist}
\usepackage[export]{adjustbox} 
\soulregister\cite7 
\soulregister\citep7 
\soulregister\citet7 
\soulregister\ref7 
\soulregister\pageref7
\soulregister\sysname7
\usepackage{pifont}
\usepackage{hyperref}
\usepackage{tikz}
\usepackage{fontawesome5}
\hypersetup{
colorlinks=true,
linkcolor=black,
}
\newcommand*\circled[1]{\tikz[baseline=(char.base)]{
            \node[shape=circle,draw,inner sep=0.5pt,fill=white,text=black] (char) {#1};}}

\newcommand{\sysname}{\textrm{\textit{\textbf{MobiFuse}}}\xspace}
\newcommand{\dataset}{\textrm{\textit{RealToF}}\xspace}
\def\BibTeX{{\rm B\kern-.05em{\sc i\kern-.025em b}\kern-.08em
    T\kern-.1667em\lower.7ex\hbox{E}\kern-.125emX}} 
\begin{document}

\title{MobiFuse: A High-Precision On-device  Depth Perception System with Multi-Data Fusion}

\author{Jinrui~Zhang,~Deyu Zhang,~\IEEEmembership{Member,~IEEE,}~Tingting Long, Wenxin~Chen, ~Ju~Ren,~\IEEEmembership{Senior Member,~IEEE,} Yunxin Liu,~\IEEEmembership{Senior~Member,~IEEE},Yudong Zhao, ~Yaoxue~Zhang,~\IEEEmembership{Senior~Member,~IEEE} and~Youngki Lee
	\IEEEcompsocitemizethanks{
		\IEEEcompsocthanksitem Jinrui Zhang, Deyu Zhang, Tingting Long, Wenxin Chen are with the School of Computer Science and Engineering, Central South University. Changsha 410083, China. E-mails:  \{zhangjinrui,zdy876, TingtingLong,224712252\}@csu.edu.cn
		\IEEEcompsocthanksitem Ju Ren and Yaoxue Zhang are with the Department of Computer Science and Technology, Tsinghua University, Beijing 100084, China. E-mails:\{ renju, zyx\}@tsinghua.edu.cn
		\IEEEcompsocthanksitem Yunxin Liu is with Institute for AI Industry Research (AIR), Tsinghua University, Beijing 100084, China. E-mails: liuyunxin@air.tsinghua.edu.cn
        \IEEEcompsocthanksitem Yudong Zhao is with SHANGHAI TRANSSION CO., LTD, Shanghai 201203, China. E-mails: yudong.zhao@transsion.com
		\IEEEcompsocthanksitem Youngki Lee is with the Department of Computer Science and Engineering, Seoul National University, Seoul 08826, Republic of Korea. E-mails: youngkilee@snu.ac.kr
		\IEEEcompsocthanksitem Deyu Zhang is the corresponding author. \\ \linebreak
	}

}

\maketitle

\begin{abstract}
We present \sysname, a high-precision depth perception system on mobile devices that combines dual RGB and Time-of-Flight (ToF) cameras. To achieve this, we leverage physical principles from various environmental factors to propose the Depth Error Indication (DEI) modality, characterizing the depth error of ToF and stereo-matching. Furthermore, we employ a progressive fusion strategy, merging geometric features from ToF and stereo depth maps with depth error features from the DEI modality to create precise depth maps. Additionally, we create a new ToF-Stereo depth dataset, \textit{RealToF}, to train and validate our model. Our experiments demonstrate that \sysname excels over baselines by significantly reducing depth measurement errors by up to 77.7\%. It also showcases strong generalization across diverse datasets and proves effectiveness in two downstream tasks: 3D reconstruction and 3D segmentation. The demo video of \sysname in real-life scenarios is available at the de-identified YouTube link \href{https://youtu.be/jy-Sp7T1LVs}{\color{red}\faYoutube}.  

\end{abstract}

\section{Introduction}\label{introduciton}
\IEEEPARstart{A}{ccurate} depth perception plays a crucial role in 3D perception on mobile devices, including depth-dependent tasks like 3D reconstruction~\cite{yang2020mobile3drecon}, 3D segmentation~\cite{zhang2023cmx}, 3D pose estimation~\cite{choi2021mobilehumanpose}, AR~\cite{zhao2022litar} and bokeh effect rendering~\cite{dutta2021depth}. 
Typical depth perception on mobile devices relies either on RGB-based stereo matching~\cite{zhang2022mobidepth, wang2018anytime} or Time-of-Flight (ToF) sensors~\cite{xie2023_mozart,tofarmobile}. Yet, both approaches exhibit shortcomings in real-life scenarios, greatly limiting their applicability. The stereo-matching method fails in areas with weak texture or uneven illumination, as it cannot identify the corresponding pixels in these areas. On the other hand, the mobile ToF sensor cannot measure depth on surfaces with specular reflection, low reflective materials, and objects aligned parallel to the optical axis~\cite{zhang2022indepth}.

Fusing depth from dual RGB cameras and ToF harnesses the complementary advantages of both approaches, i.e., ToF sensors perform well in weak-texture areas, while stereo-matching methods are highly robust to low object reflectance. However, certain existing approaches~\cite{agresti2017deep, marin2016reliable} relying on using obtained ToF confidence scores to perform a weighted average of ToF and stereo matching depth values, which introduces two limitations: (1) Simply merging the cross-modality data using interpolation. They overlooked the substantial differences between the multi-modal data obtained from ToF and stereo cameras. It leads to the loss of modality-specific features, which is essential for ToF-Stereo fusion. (2) Inadequate ToF depth confidence estimation. They only consider signal amplitude for estimation, which proves inaccurate for depth error prediction (Sec. \ref{subsec_motivation}). 

To address all the above issues and enhance the final depth accuracy, we propose \sysname, an efficient on-device depth perception system that harnesses a synergy of data from prevalent mobile ToF and dual cameras on off-the-shelf mobile devices. The system analyzes and constructs physical relationships between environmental factors and the depth errors in ToF and stereo matching. Fusing these relationship features with ToF and stereo-matching depth information produces accurate fused depths.  
 
However, designing the system faces two key challenges:

\begin{inparaenum}[1)]
\item \textit{Difficulty in precisely and efficiently comparing better depth measurement between mobile ToF and stereo matching.} This challenge originates from their fundamentally different depth perception principles, leading to distinct depth error profiles. ToF is sensitive to objects' materials and surface characteristics, while stereo matching is more influenced by object texture distribution and lighting conditions. Additionally, achieving accurate comparison across heterogeneous measurements requires complex models and heavy computation~\cite{pang2020clocs}, which can result in unacceptable latency for resource-constrained mobile devices. To make matters worse, no reliable detectors exist to effectively assess the depth errors specifically for mobile ToF. 

\item \textit{Absence of a real-world dataset of ToF-Stereo depth perception.} 
In order to facilitate optimal ToF-Stereo fusion, it is crucial to thoroughly capture the factors that influence depth measurements of mobile ToF and stereo matching within real-world scenarios. This task entails collecting pixel-aligned data with accurate ground truth across diverse scenes and multiple devices, which presents significant challenges. Existing ToF-Stereo datasets are quite limited in terms of scene diversity, scale, and reality, as depicted in Table~\ref{Table_dataset}. ToF-100~\cite{gao2021joint} and ToF-18k~\cite{zhang2018deepdepth} datasets only contain ToF data. SYNTH3~\cite{agresti2017deep} is a synthetic dataset comprising 55 samples. These data are generated by a 3D rendering engine. However, it cannot fully replicate the complexity of real-world environments, such as ambient light and scene structure. Although the REAL3~\cite{dal2015probabilistic} and LTTM5~\cite{agresti2019stereo} have both ToF and stereo data in real-world scenes, they are limited to only 8 and 5 samples, respectively. Furthermore, their data lacks precise pixel-level alignment.  
\end{inparaenum}

\begin{table}[t!]
\centering
\caption{Comparison between our \textbf{\textit{RealToF}} dataset and other ToF-related datasets.}

\setlength{\tabcolsep}{1.6mm}{
\begin{tabular}{|c|c|c|c|c|}
\hline
\rowcolor[HTML]{EFEFEF} 
\textbf{Dataset} & \textbf{Scenes}  & \textbf{Reality} & \textbf{ToF\&Stereo} & \textbf{Pixel Alignment}\\ \hline
ToF-100~\cite{gao2021joint} & 100 &  
\CheckmarkBold & 
\XSolidBrush &
\XSolidBrush   
\\ \hline
ToF18k~\cite{zhang2022indepth} &  150&  
\CheckmarkBold & 
\XSolidBrush &
\XSolidBrush   
\\ \hline
SYNTH3~\cite{agresti2017deep} & 55 &  
\XSolidBrush & 
\CheckmarkBold  &
\CheckmarkBold   
\\ \hline
REAL3~\cite{agresti2019stereo} & 8 &  
\CheckmarkBold & 
\CheckmarkBold &
\XSolidBrush  
\\ \hline
LTTM5~\cite{dal2015probabilistic} & 5 &  
\CheckmarkBold & 
\CheckmarkBold &
\XSolidBrush   
\\ \hline
\rowcolor[HTML]{FCEBD2}
\textbf{RealToF} & \textbf{150} &  
\CheckmarkBold  & 
\CheckmarkBold  &
\CheckmarkBold   
\\ \hline
\end{tabular}}
\label{Table_dataset}
\end{table}

To overcome these issues, we develop TSFuseNet, an end-to-end lightweight multi-data fusion model, tailored to mobile ToF and resource-constrained mobile devices. TSFuseNet utilizes the progressive fusion framework in a two-stage training process. We first analyze the physical principles connecting environmental factors to depth and stereo-matching errors. These relationships are formulated into specific loss functions, enabling the model to learn unique feature representations for each factor and map them to depth errors in ToF and stereo-matching. Consequently, the model acquires the Depth Error Indication (DEI) feature. Then, it performs cross-modal fusion by combining these features with depth information and using backward connections. This approach generates a precise depth map that blends high-level semantic and low-level geometric details.

In the design of TSFuseNet, we integrate physical principles into the model learning process. These principles act as additional constraints to guide the model, reducing its complexity and enhancing its ability to learn more effective and generalizable representations for new data. Furthermore, we employ a progressive fusion strategy: (i) fusing the DEI features of ToF and stereo-matching with their depth information, and (ii) utilizing a backward-connected mechanism to facilitate bidirectional information flow across the two modalities, enabling the learning of inter-modal correlations. This integration combines geometric features from the depth modality of ToF and stereo-matching with depth error features from the DEI modality, generating a precise depth map. 


For training and validating our model, we create a diverse-scene ToF-Stereo depth dataset with pervasive ground truth in the real world, namely \textit{RealToF}, to substantiate our model designs and quantitatively assess the efficacy of \sysname. The dataset consists of data collected from 150 diverse indoor scenes using the ToF and dual RGB cameras on the Huawei P40Pro. We build a compact and stable multi-device acquisition rig for facilitating data collection in various scenarios. To establish accurate ground truth, we jointly use the Intel RealSense depth camera~\cite{RealSense} and a laser rangefinder~\cite{deli} to improve the precision of the ground truth.

We implement \sysname on several commodity mobile devices with ToF and conduct comprehensive experiments. Evaluation results show that \sysname significantly outperforms prior depth completion methods, e.g., GuideNet~\cite{tang2020learning} and InDepth~\cite{zhang2022indepth}, and depth fusion methods, TSFusion~\cite{agresti2017deep}, in terms of accuracy, latency, and energy consumption. Compared with GuideNet, InDepth and TSFusion, \sysname~achieves a remarkable reduction of 77.7\%, 40.3\%, and 59.5\%in mean absolute error(MAE), respectively, and has reduced latency by 84.4\%, 81.6\% and 6.8\%, reduced energy consumption by 89.2\%, 85.6\% and 9.9\% on Huawei P40Pro. We also conduct rigorous evaluations of the system across various datasets, and the results indicate that MobiFuse outperforms other systems in terms of its generalization capabilities. Furthermore, we evaluate the accuracy of using depth results obtained from \sysname compared to other methods in depth-based 3D reconstruction and 3D segmentation. The results show that both \sysname-based 3D reconstruction and 3D segmentation achieve the best performance.



In summary, the main contributions are as follows:
\begin{itemize}
    \item We propose \sysname, an end-to-end ToF-Stereo fusion system for accurate depth perception in diverse scenes that run efficiently on commodity mobile devices.
    \item We propose a new DEI modal information by considering multiple real-life environmental factors with physical principles to characterize the depth error of ToF and Stereo accurately. Furthermore, we employ a progressive fusion approach with a backward connection to merge the geometric features from the depth modality with the depth error features from the DEI modality, generating accurate fused depth maps.
    \item We create the \textit{RealToF}, a real-life mobile ToF-Stereo depth dataset with precision ground truth in diverse scenes. We plan to open-source the dataset and provide ongoing updates upon paper acceptance. 
    \item We apply \sysname to two downstream 3D applications, confirming its ability to significantly enhance the performance of these applications. 
\end{itemize}
\section{Background and motivation}
\label{sec_background}
\subsection{Principle of ToF measurement}
\begin{figure}[t]
	\centering
	\includegraphics[width=1\linewidth]{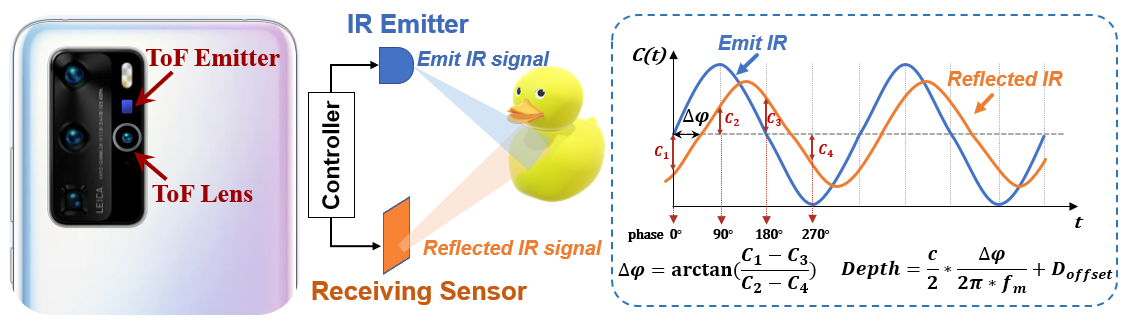}
	\caption{Simplified illustration of ToF measurement.} 
	\label{ToF_principle}
\end{figure}

\textbf{Mobile ToF depth measurement.} The mobile ToF sensor emits infrared (IR) light to illuminate the scene and captures the reflected IR light from objects, as shown in Figure~\ref{ToF_principle}. The existing ToF sensors can be cast into dToF and iToF. dToF measures the depth by directly measuring the round-trip time($t$) of the infrared signal between emission and reflection. iToF derives the depth from the phase offset ($\Delta \varphi $) in emitted and received continuous wave infrared signals. Most commodity mobile devices (especially Android smartphones) adopt iToF due to its low cost and relatively high resolution~\cite{DayDayNews}, e.g., Huawei P40Pro, Samsung S20+, etc. Mobile ToF in this paper refers to iToF unless otherwise indicated. 

To obtain a depth map, the mobile ToF gathers the samples, denoted as $C_{1},C_{2},C_{3}$, and $C_{4}$ at four phase angles. Based on the samples, mobile ToF determines the phase offset $\Delta \varphi$, and then calculates the object distance as shown in Figure~\ref{ToF_principle}, where $c$ denotes the light speed, $f_{m}$ is the signal modulation frequency and $D_{offset}$ is a depth offset coefficient.

\subsection{Motivation Study}
\label{subsec_motivation}
\textbf{Advantages in depth complementarity between mobile ToF and stereo matching.} 
Fusing depth maps from mobile ToF and stereo matching can fully utilize the complementary advantages of both methods. 
To explore the characteristics of the two depth measurement methods, we obtain over 100 sets of depth maps in multiple scenes, using both mobile ToF and stereo matching. Then we analyze the depth error distribution of all pixels, as shown in Figure 2(a). 
In the figure, we divide four regions based on a depth error threshold of 60mm\footnote{Depth errors in this paper refer to absolute error unless otherwise indicated.}, considering the errors beyond this limit as large ones. Based on our statistics, region \circled{1} and \circled{4} account for 15\% and 20\% of the pixels, respectively. In these regions, either the ToF or stereo matching achieves accurate depth measurement. 
61\% of the pixels fall into region \circled{3}, where both ToF and stereo matching demonstrate small depth errors. Hence, by fusing the depth of ToF and stereo matching across these 96\% of pixels, we can consistently select high-precision depth values from either method. For only 4\% of the pixels(in region \circled{2}),  neither ToF nor stereo matching can provide satisfactory depth measurement. These results show that we can achieve accurate depth measurement in most cases(96\% pixels), which proves the great potential of ToF-Stereo fusion for mobile depth perception.

\begin{figure}[h!]
    \centering
    \includegraphics[width=1\linewidth]{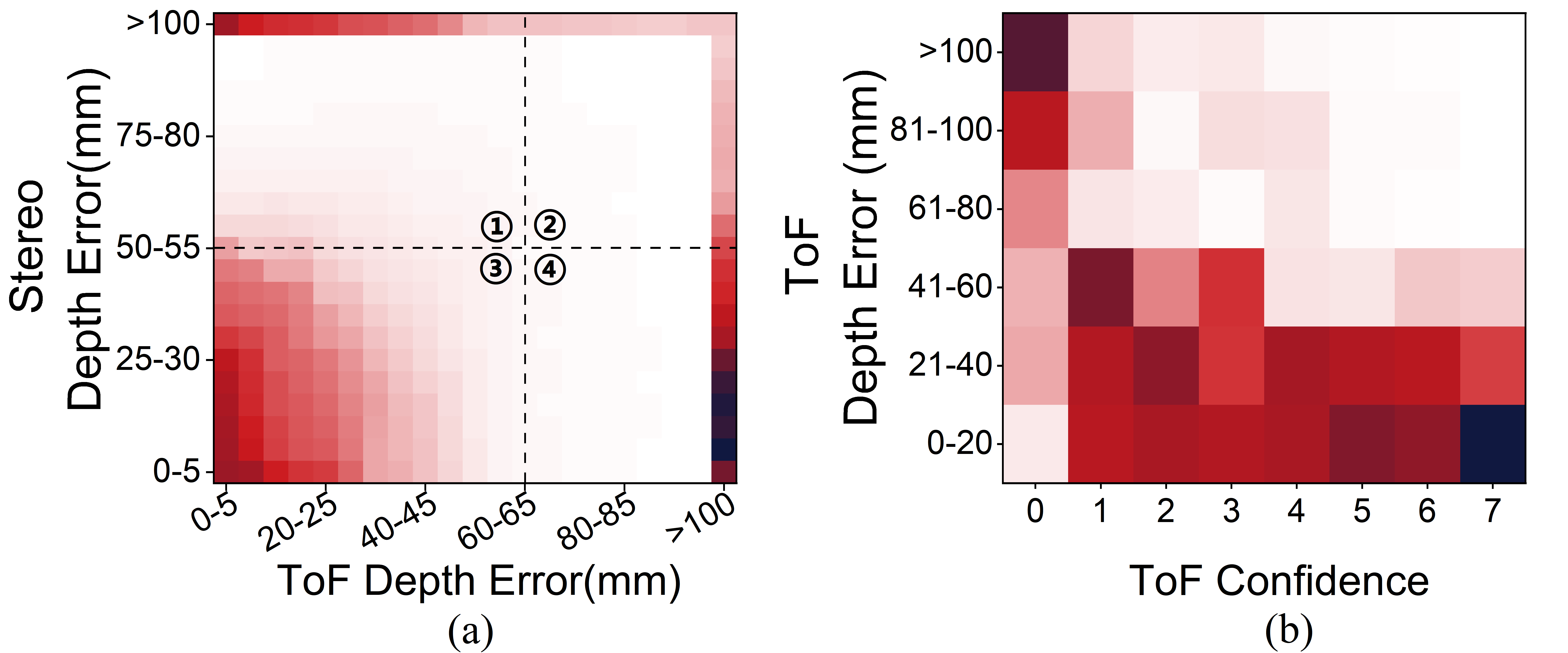}
    \caption{(a) Depth error distribution of mobile ToF and stereo-matching for each pixel. (b)  Correlation of existing mobile ToF confidence and depth error. Darker colors indicate more pixel counts.}
    \label{d-e-c}
\end{figure}

\textbf{Existing signal amplitude-based mobile ToF confidence cannot assess the depth error precisely.} 
Mobile ToF usually provides confidence values for each depth pixel in depth maps. Generally speaking, this kind of confidence is based on the amplitude of received IR signals. Our experimental findings unveil that the existing confidence of ToF fails to represent the veritable extent of depth errors. As shown in Figure 2(b), we collect 500 depth images with two ToF-enabled Android smartphones, e.g., Huawei P40Pro and Samsung S20+, and find that the ToF confidence level\footnote{The confidence on the Android mobile device is evenly divided into 8 levels(0-7), where a higher number denotes higher confidence.}  only statistically correlates with the probability distribution of depth errors, which cannot precisely assess the depth error. Specifically, lower confidence levels are associated with a higher likelihood of pixels with significant errors, and vice versa. However, even with the lowest confidence, the error of a pixel could be as small as 10mm, while with the highest confidence, the error could be as high as 60mm. 
Consequently, the existing mobile ToF confidence cannot serve as a precise indicator to compare the accuracy of ToF and stereo matching in terms of depth error.

\section{MobiFuse Overview}
\label{sec_MobiFuse_Overview}
\subsection{Design Goals} 
\textbf{High-precision depth.} Our primary target is to synergistically complement the individual strengths of ToF and stereo matching in depth measurement, enabling the generation of highly accurate depth maps. To do so, we need to compare the depth errors of ToF and stereo for each pixel, selecting the depth with the minimum error as the definitive outcome.

\noindent \textbf{Fully end-to-end mobile system.} We strive to develop an effortless plug-and-play system on resource-constrained mobile devices. Diverging from existing methods like InDepth~\cite{zhang2022indepth}, our approach eliminates the need for offloading data to external servers, ensuring data privacy and avoiding system instability in unreliable network environments.

\noindent \textbf{Reduced runtime latency.} As the system consists of multiple modules, we deploy various modules across heterogeneous computing units to minimize the end-to-end latency. 

\subsection{System Architecture}
\begin{figure*}[h!]
	\centering
	\includegraphics[width=1\linewidth]{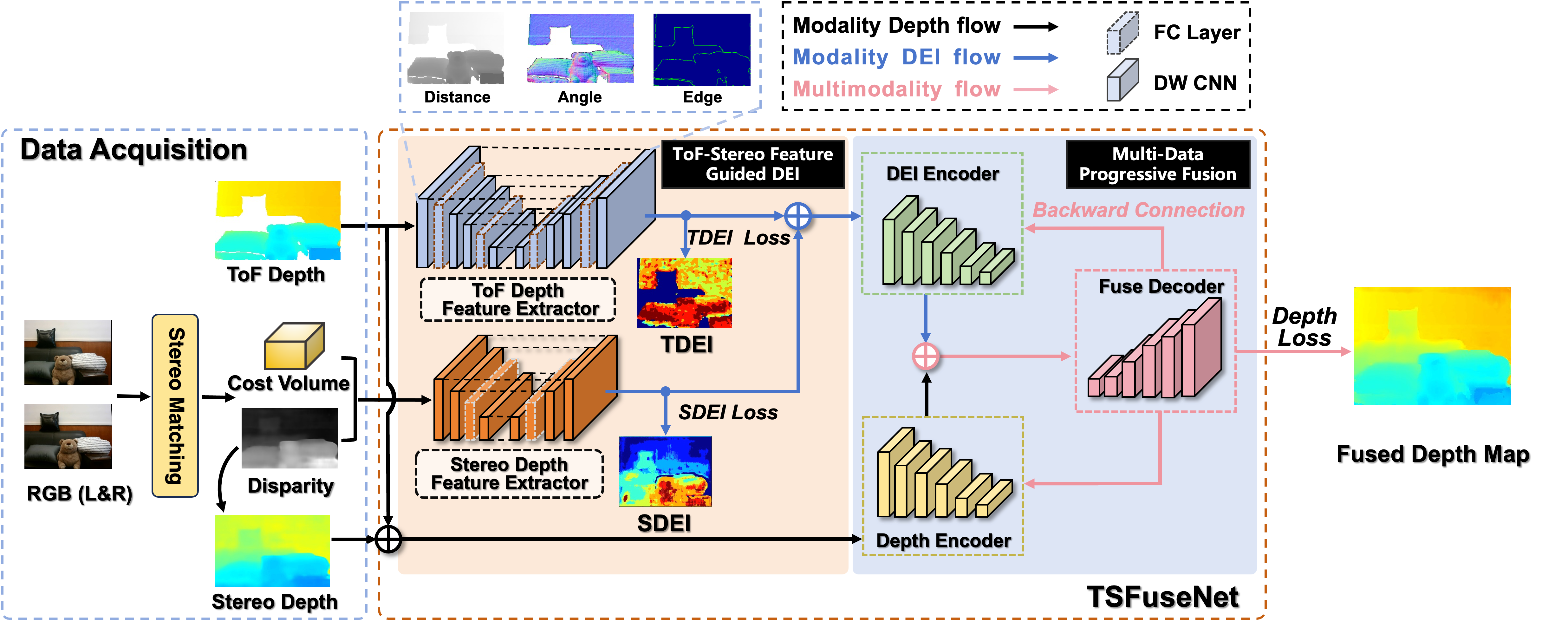}
	\caption{The system architecture of \sysname.} 
	\label{pipeline}
\end{figure*}

Figure~\ref{pipeline} illustrates the pipeline of \sysname, which contains two components: data acquisition and the TSFuseNet model. 
In the data acquisition component,  we use a stereo-matching algorithm, i.e, we employ our lightweight optimized MSNet~\cite{shamsafar2022mobilestereonet} as the backbone, to extract the cost volume, disparity, and stereo depth map from paired RGB images, which other stereo matching algorithms can flexibly substitute. These data pieces, along with the depth map acquired by mobile ToF, collectively serve as input to the model. 

The TSFuseNet model comprises two key modules: the ToF-Stereo Feature Guided Depth Error Indication (DEI) module and the Multi-Data Progressive Fusion module. The DEI module uses a dual-branch structure to create DEI networks for ToF and stereo-matching by incorporating environmental factors and physical principles. This equips the model with the ability to indicate depth errors. In the fusion module, we use the progressive fusion approach to align and merge DEI features from ToF and stereo-matching with raw depth map features. Then, we adopt a backward connection operation that facilitates bidirectional information exchange between ToF and stereo-matching modalities within the network, integrating geometric and error features from the depth and DEI modalities separately to generate a precise fused depth map.

\section{Model Design} 
\label{TSFuseNet}
Current ToF-stereo fusion methods inaccurately characterize pixel-wise depth errors due to the sole reliance on amplitude intensity for ToF confidence. Additionally, the lack of consideration for correlated information between ToF and stereo-matching data during depth map merging leads to limited accuracy in the fused depth maps.

To tackle these issues, we conduct preliminary experiments to identify environmental factors affecting mobile ToF depth errors. By analyzing the physical principles between these factors and the errors, we guide the design of the model structure and loss function before model development.

\subsection{Demystify Depth Error of Mobile ToF}
\label{sec_demystify}
We have analyzed 2500 images captured with mobile ToF to pinpoint the primary factors influencing ToF depth error.

\begin{figure*}[htb]
    \centering     
    \includegraphics[width=1.0\linewidth]{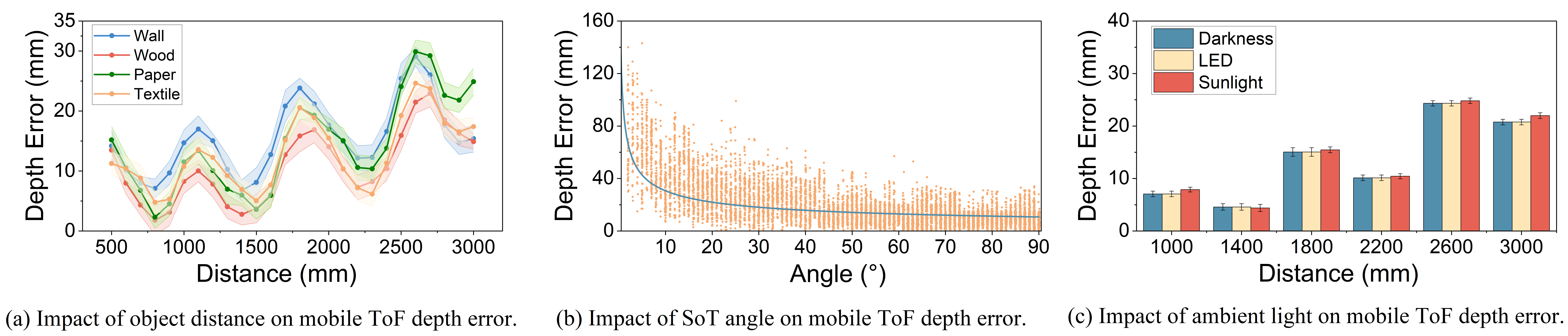} 
    \caption{Influence of different factors on mobile ToF depth error in real-life scenarios.}
    \label{observation}
\end{figure*}

\textbf{1) The depth error increases with the object distance, following a sinusoidal pattern rather than a linear one.} 
To explore the correlation between mobile ToF depth error and distance, we oriented an object's surface perpendicular to the ToF camera at varying vertical distances. Our findings reveal that the depth error not only escalates with object distance due to path loss but also exhibits a sinusoidal pattern, as shown in Figure~4(a). This trend persists across different material types and mobile ToF models (Samsung S20+, Huawei Mate30Pro).

The sinusoidal pattern emerges from the square wave emission of practical mobile ToF systems, introducing harmonic errors that cause oscillating deviations in distance measurements~\cite{rapp2007experimental}. 
The correction of harmonic errors is out of the scope of our work, thus, we need to model the relationships between object distance and depth error in this pattern.

\textbf{2) Significant impact of the surface-to-ToF optical axis(SoT) angle\footnote{The SoT angle refers to the angle between the optical axis of the mobile ToF and the surface of an object.} on depth errors, especially in the small angle region.} We select all pixels with the ground truth depth at $0.8 m$ in the \textit{RealToF} dataset, and calculate the corresponding SoT angle~\cite{depthtonormal}, as shown in Figure~4(b)\footnote{$0^\circ$ and $90^\circ$ indicate parallel and perpendicular alignments, respectively.}.  By averaging the depth errors concerning the angles, we find that the error decreases exponentially. Within a small angle region, the error decreases significantly with an increasing angle. Then the decrease slows down with larger angles. The smallest error is achieved at the object surface orthogonal to the optical axis of the mobile ToF. The reason is that the intensity of the reflected infrared signal decreases with the SoT angle. Therefore, a normal map, which is a type of texture map that encodes SoT angles of an object, is necessary for assessing depth errors of mobile ToF.

\textbf{3) Object edges bring higher depth error.} In our preliminary experiments, 80\% of pixels at object edges show depth loss in ToF, with 15\% of pixels having depth errors over 80mm. This is due to the integration of light paths from the foreground object and background over the apertures, causing significant depth measurement errors at object edges in mobile ToF~\cite{chugunov2021mask}. Hence, we need to pay more attention to the depth measurement of object edge pixels. 

\textbf{4) Adjacent pixels on a plane with the same material have close depth errors.} According to the surface normal constraint~\cite{long2021adaptive}, neighboring pixels on the same plane with identical material tend to have continuous depth values and close depth errors. This is also known as local consistency. We can use this constraint to aid us in assessing depth error.

\textbf{5) Negligible impact of ambient light on mobile ToF depth error.} We conducted experiments in dark, LED light, and sunlight environments, respectively. 

Firstly, we find that the LED light has no impact on the mobile ToF depth measurement. The reason is that the typical wavelengths of mobile ToF ($850/940nm$) and LED ($380-780nm$) do not overlap. Furthermore, a spectral filter has been integrated into the ToF lens to filter out light noise, minimizing interference at the $940 nm$ wavelength. 

Secondly, in the effective range, the sunlight only slightly impacts the mobile ToF depth error, i.e., compared to darkness testing, the maximum absolute error increased by only 2mm, as shown in Figure~4(c). There are two main reasons for this: (a) Continuous-wave iToF cameras collect multiple samples per measurement and adjust the phase shift by subtracting energy samples, effectively reducing the impact of sunlight-induced energy offset. (b) Atmospheric water molecules greatly absorb the $940 nm$ infrared light in sunlight before it reaches the ground~\cite{Sunlight}, diminishing its intensity and minimizing its impact on mobile ToF depth accuracy. 



\textbf{Summary.} Based on the analysis above, we identify the dominant factors to model the mobile ToF errors as object distance, SoT angles, and scene structures, e.g., object edges and planes. All the dominant factors are conveyed in the ToF depth map simply in runtime.

\subsection{TSFuseNet Architecture}\label{subsec_tdei} 
The TSFuseNet is designed with two modules: the ToF-Stereo Feature Guided DEI module and the Multi-data Progressive Fusion module. The DEI module integrates a physics-driven loss function based on preliminary experiments to provide a structured understanding of the data, simplifying the model's complexity and enhancing its generalization. In the fusion module, we employ a progressive fusion strategy, merging distinct data features from ToF and stereo-matching depth maps and depth error representations. To enable bidirectional information transfer between different modalities within the network, we employ a backward-connected mechanism. This approach aids the model in capturing correlated information across modalities, enhancing its depth fusion capabilities.

\subsubsection{ToF-Stereo Feature Guided DEI Module}
In DEI module, we adopt a dual-branch architecture, e.g., ToF branch and Stereo branch, to initially establish the complex relationship between environmental factors and ToF/stereo depth errors. 

\textbf{ToF branch.} This branch takes ToF depth maps as input and uses a U-Net structure-based\cite{ronneberger2015u} depth encoder-decoder to extract ToF depth features, capturing critical aspects like distance, angle, object edges, and planes correlated with ToF depth errors. Our feature extractor comprises nested $3\times3$ depthwise convolutional(DW CNN) layers and fully connected(FC) layers, incorporating up-sample, down-sample, and skip connections. The DW CNN is used to capture local spatial information from object edges, planes, and other features, which are then combined globally through FC layers in a nonlinear manner and mapped to the output categories. By extracting the output features of the ToF branch, we derive the intermediate result termed \textbf{T-DEI}. This parameter can serve as an online indicator of the magnitude of the ToF depth error.

\textbf{TDEI loss function.} To integrate the physical principles between ToF depth errors and environmental factors into the model, we design the TDEI Loss ($\mathcal{L}_{TDEI} $), which comprises two components, i.e., Edge-aware DEI Loss ($\hat{\mathcal{L}}_{DEI}$) and Local Consistency Loss ($\mathcal{L}_{Lc}$).
\begin{equation}
   \mathcal{L}_{TDEI}=\alpha \hat{\mathcal{L}}_{DEI} +\beta \mathcal{L}_{Lc}
\end{equation}
where $\alpha, \beta$ are the coefficients that weigh the contribution of each component. 

\textbf{\small (a) Edge-aware DEI Loss.}  This loss function guides the module training to minimize the gap between the predicted and ground truth ToF DEI levels. By doing so, it thoroughly considers all the dominant factors identified in Section \ref{sec_demystify}, e.g., object distance and angles. Furthermore, we locate the object edges according to the gradient of pixels in the ToF depth map. Then assign a higher weight to the pixels on the object edge, such that the module training pays more attention to the prediction of DEI level on object edges.
\begin{equation}
    \hat{\mathcal{L}}_{DEI} =-\sum_{i=0}^{N}\sum_{k=0}^{K}exp(-|\triangledown E_{i}|)y_{i,k}log(p_{i,k})
\end{equation}
where $N$ represents the number of pixels, $K$ denotes the number of DEI levels. $y_{i,k}$ is the true label of the $k-th$ level for $i-th$ pixel, and $p_{i,k}$ represents the predicted probability of the $k-th$ level for $i-th$ pixel. $\triangledown E_{i}$ is the depth value gradient of $i-th$ pixel.

\textbf{\small (b) Local Consistency Loss.} We discriminate the neighboring pixels on the same plane according to the variance of their SoT angle from normal map~\cite{long2021adaptive}. These pixels tend to have continuous depth measurements and close depth errors. Hence, the depth error variation of pixels on the same plane with identical textures exhibits consistency with the changes in the pixel's surface SoT angle. To capture this, we assign a higher weight to the neighboring pixels on the same plane but with a larger gap in predicted errors:
\begin{equation}
    \mathcal{L}_{Lc}=\frac{1}{|N|}\sum_{i=0}^{N}log(|\sigma_{i}^{P}-\sigma_{i}^{n}|)*\left \| \sigma _{i}^{P}-\sigma _{i}^{GT} \right \|^{2}
\end{equation}
We compute the variances of the predicted ToF \textit{DEI} and the ground truth \textit{DEI} for each pixel concerning its eight neighboring pixels, denoted as $\sigma _{i}^{P}$ and $\sigma _{i}^{GT}$, respectively. $\sigma_{i}^{n}$ represents the variance of the surface SoT angles between each pixel and its neighboring pixels. $N$ is the total number of pixels in the dataset.

\textbf{Stereo branch.} Prior studies like MSM and CUR~\cite{hu2012quantitative} confirm the impact of cost volume and disparity quality on stereo-matching depth errors. Yet, these methods only account for a single modal factor, making them prone to image noise and suffer from poor accuracy in practical scenarios. To this end, we take both cost volume and disparity as inputs in the Stereo branch to investigate their correlation with stereo-matching depth error. We adopt a similar feature extractor as the ToF branch, albeit lighter due to the relatively straightforward correlation between the cost volume, disparity, and stereo-matching depth error. Also, the SDEI Loss uses a simple but effective multi-class cross-entropy loss. The Stereo branch also yields an intermediate result, termed \textbf{S-DEI}, representing the magnitude of stereo-matching depth error.

\textbf{DEI label.} To accurately capture the relationship between the factors and depth error for both ToF and stereo-matching, we introduce the DEI label. The DEI labels correspond to depth error values. An excessive amount of categories can lead to sparse data and hinder the model's ability to learn category-specific features, impacting training convergence and overall performance. Conversely, too few categories limit the model's grasp of finer details, confining its learning to general patterns and hindering accurate pattern recognition.  Therefore, we analyze the distribution of pixel counts with different depth error values in the dataset and categorize the DEI labels into 8 levels (0 to 7), ensuring a balanced data distribution upon classification. The specific correspondence between DEI levels and depth errors is presented in Table~\ref{DEI Level}. 

\begin{table}[h!]
\centering
\caption{Relation between DEI Levels and Depth Errors.}
\begin{threeparttable}
\setlength{\tabcolsep}{1mm}{
\begin{tabular}{ccccccccc}
\toprule[1.0pt]
DEI Levels         & 7            & 6         & 5          & 4          & 3          & 2          & 1           & 0                 \\ \hline
 Errors(mm) & \textless{}5 & 5$\sim$15 & 15$\sim$25 & 25$\sim$40 & 40$\sim$60 & 60$\sim$80 & 80$\sim$100 & \textgreater{}100 \\
 \bottomrule[1.0pt]
\end{tabular}}
\end{threeparttable}
\label{DEI Level}
\end{table}

\subsubsection{Multi-data Progressive Fusion Module}
In the fusion module, we use the progressive fusion strategy to incrementally combine DEI features and raw depth data from ToF and stereo-matching for more accurate depth fusion.  We combine TDEI and SDEI features from the DEI modules using the DEI Encoder to enhance the representation of ToF and stereo-matching depth errors. Yet, the fused DEI features lack certain low-level geometric details lost during network propagation, like depth gradients, divergence, and disparity curl. To address this, we use a Depth Encoder to extract these details from the raw depth data. These geometric features, along with the merged DEI features, are fed into the Fusion Decoder to integrate high-level semantic and low-level geometric information. To align features across Depth and DEI effectively, we apply a backward-connected strategy, feeding deep features from the Fusion Decoder back into the Depth and DEI Encoders. This feedback refines the feature extraction process by guiding it with high-level information from cross-modal fusion, promoting deep cross-modal information fusion during feature extraction.

\textbf{Depth loss function.} We train the overall fused depth of the model using a Depth Loss function $\mathcal{L}_{depth}$. The final depth loss function includes an $L1$ norm error term and a smoothing regularization term, as shown below:
\begin{equation}
    \mathcal{L}_{depth}=\frac{1}{n}\sum_{i=1}^{n}|R_i|+\mu\;\frac{1}{n}\sum_{i}(|\triangledown_xR_i|+|\triangledown_yR_i|)
\end{equation}
where $R_i=\hat{I}_{i}^{depth}-I_{i}^{depth}$, $\hat{I}_{i}^{depth}$ is the estimated depth, $I_{i}^depth$ is the depth ground truth. $\triangledown_x$ and $\triangledown_y$ are the spatial derivatives for the x-axis and y-axis.

Utilizing an $L1$ norm error term enables capturing significant changes and edge information in the depth map while adapting well to background sparsity. Introducing a smoothing regularization term helps eliminate low-amplitude structures in the depth map, sharpening the main edges\cite{xian2020structure}.

\subsection{Model Training}
TSFuseNet follows a two-stage training strategy. Initially, the fusion module is frozen, allowing focused training of the ToF and Stereo branches within the DEI module to learn the relationship between environmental factors and depth errors, mapping them to T-DEI and S-DEI. These features guide the subsequent fusion step and prevent the model from converging to suboptimal solutions due to circular dependencies. In the second stage, the fusion module and both DEI branches are jointly trained to integrate ToF and stereo-matching DEI features with raw depth data. This enables the model to learn how to complement depth fusion based on the DEI information, leading to more precise depth values. Throughout this process, the Ground Truth Depth supervises the DEI module in reverse, further enhancing the ToF and stereo-matching depth error representation and improving overall depth fusion outcomes.

\section{RealToF Dataset}
\label{dataset}
To effectively design \sysname, we collect \dataset\footnote{The \textit{RealToF} dataset is available at \href{https://drive.google.com/uc?export=download&id=11T3ILpWRbN-ekevJ5UOvzkGZXIwVPJlc}{\color{blue} Google Drive.}}, a high-quality ToF-Stereo depth dataset in real-world scenes. \dataset includes 2500 meticulously curated image pairs, providing insightful observations of approximately 150 diverse indoor environments. Each image pair in the dataset provides essential data modalities, such as RGB, Depth ground truth, Stereo Depth map, Mobile ToF Depth map, and a Normal map generated from the ToF depth, as depicted in Figure~\ref{dataset}. 

To ensure the validity of the dataset, we set several goals for data collection:
\begin{inparaenum}[1)]
\item \textbf{Scene diversity.} Data collection in multiple scenarios improves model accuracy and generalization by increasing the diversity of data features and reducing bias.
\item \textbf{Precise ground truth.} The ground truth of the dataset serves as the standard for model training, and it is essential to ensure its accuracy to train accurate and reliable models.
\item \textbf{Multi images pixel-level alignment.} 
Pixel-level alignment reduces region occlusion and pixel displacement in images. As such, we can guarantee that all pixels have accurate ground truth, avoiding information loss in the dataset.
\end{inparaenum}
\begin{figure}[t]
    \centering
    \includegraphics[width=1\linewidth]{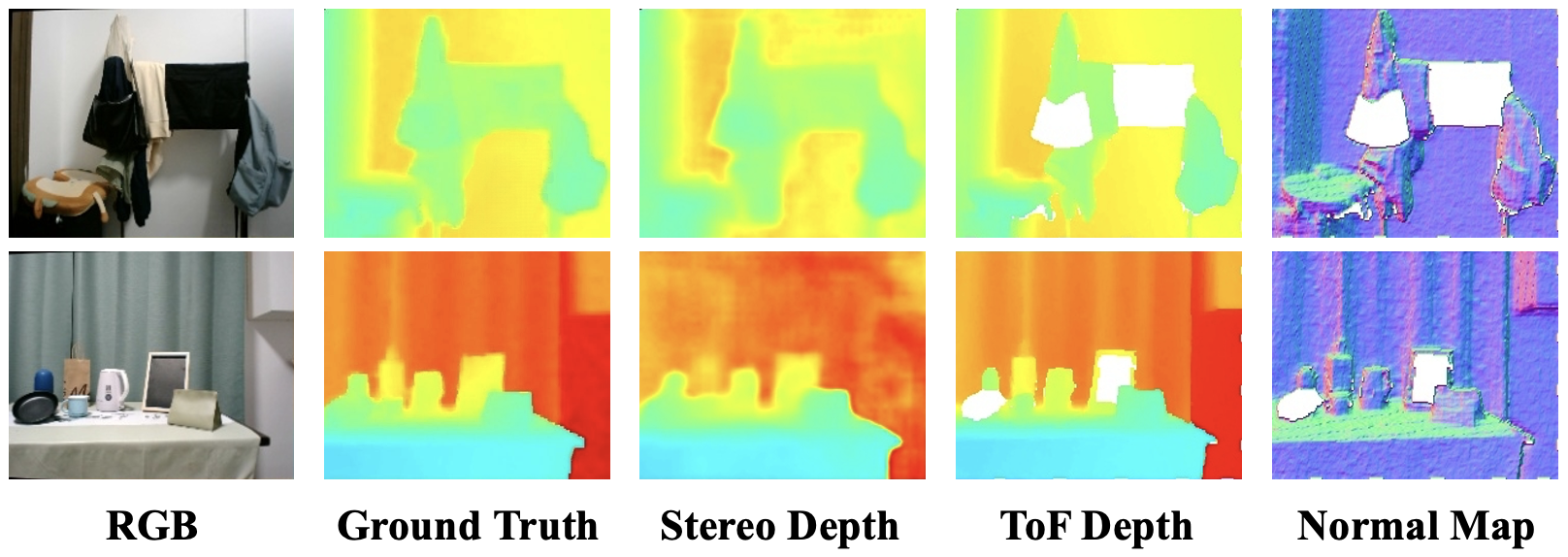} 
    \caption{Examples in \textit{RealToF} dataset.}
    \label{dataset}
\end{figure}

\begin{figure}[t]
    \centering
    \subfloat[ Dataset collection bracket.]{\includegraphics[width=0.6\columnwidth]{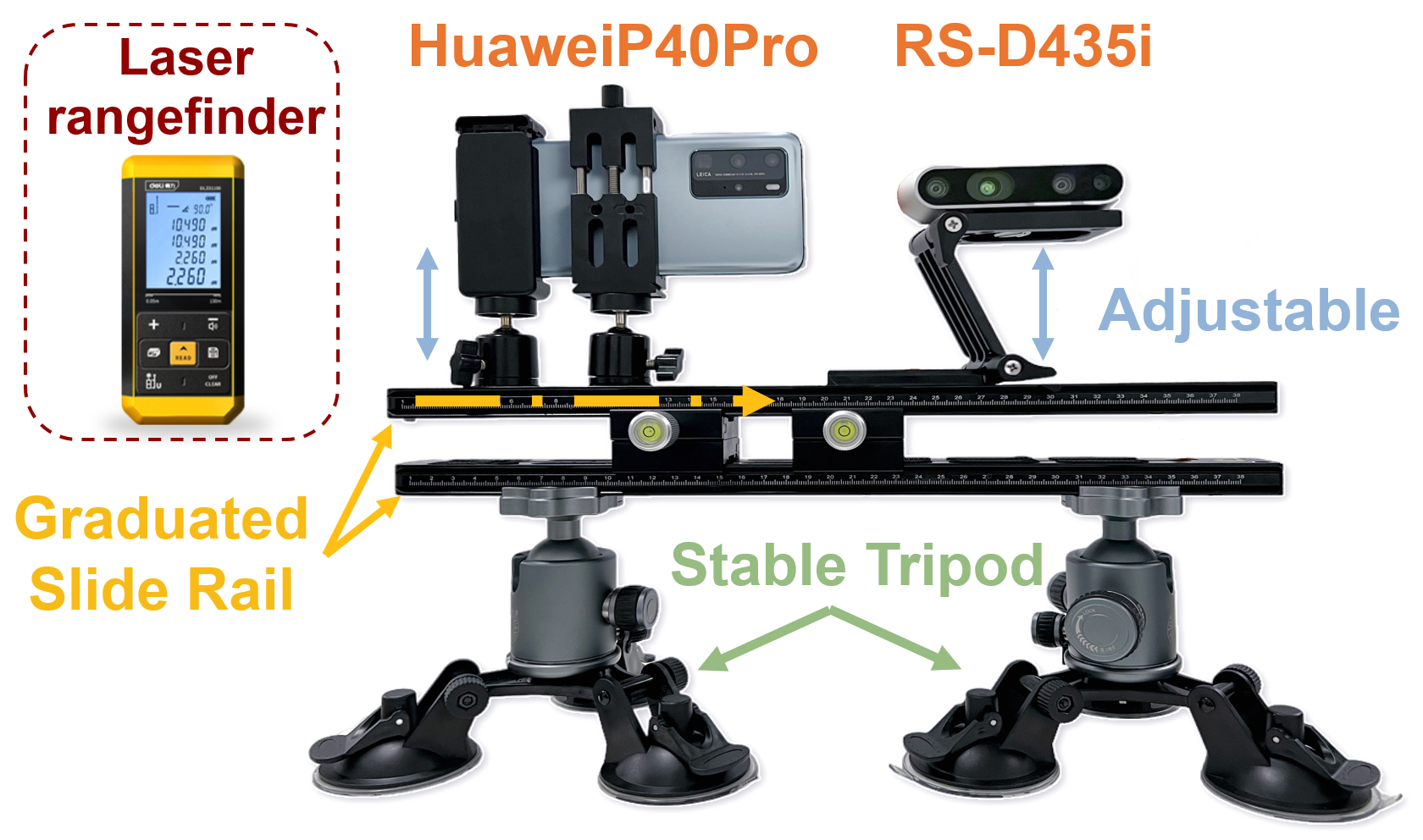} \label{data_device}}\hspace{30pt}
    \subfloat[Calibration board.]{\includegraphics[width=0.3\columnwidth]{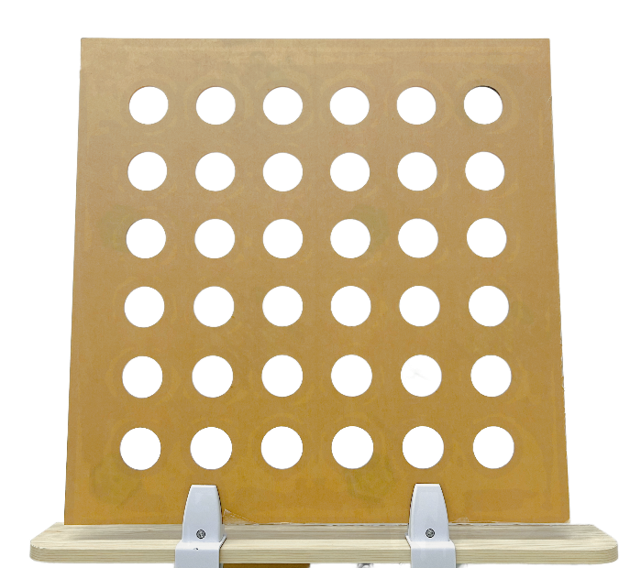} \label{2.5D pattern}}
    \caption{Hardware equipment for \textit{RealToF}  collection.}
    \label{RealToF hardware}
\end{figure}

\textbf{Multi-scene data collection.} To facilitate data collection in various scenarios,  we assemble the devices and multiple sliders on a portable bracket for data collection, as shown in Figure~\ref{data_device}. We use the Huawei P40 Pro~\cite{Huawei} to collect data for ToF and stereo depth, and two stable tripods with cushion pads to ensure its balance. It is remarkably lightweight, rendering it immensely convenient for acquiring data across diverse contexts and scenarios. 

To ensure dataset authenticity, our dataset comprises 150 diverse indoor environments, as shown in Table~\ref{tab_scene}. Our chosen scenarios cover densely inhabited residential areas, known for their frequent usage of AR/VR and 3D reconstruction applications on mobile devices. Besides, we collect 100 commonly found objects, including various materials like plastics, wood, paper, fabrics, ceramics, rubber, and metals. This diverse selection reflects the wide array of materials we encounter in our everyday lives. 

\begin{table}[h]
\centering
\caption{Scene categorization in \textit{RealToF}.}
\label{tab_scene}
\begin{threeparttable}
\setlength{\tabcolsep}{0.5mm}{
\begin{tabular}{c|ccccc}
\toprule[1.0pt]
Categories  & Office  & Meeting room & Store room  & Dormitory & Corridor \\ \hline
Scene Qty  & 50      & 20           & 10          & 15        & 5        \\
Image Qty & 500     & 300          & 300         & 300       & 150      \\ \hline\hline
Categories  & Bedroom & Living room  & Dining room & Bathroom  & Kitchen   \\ \hline
Scene Qty  & 10      & 20           & 10          & 5         & 5        \\
Image Qty & 300     & 300          & 200         & 50        & 100            \\ 
 \bottomrule[1.0pt]
\end{tabular}}
\end{threeparttable}
\label{scene}
\end{table}


\textbf{Persuasive ground truth}
We jointly use the Intel RealSense D435i(RS-D435i)~\cite{RealSense} and a laser rangefinder \cite{deli} to capture the ground truth depth. While RS-D435i boasts an impressive depth error of less than 2\% within a 2-meter range, it still has limitations in capturing object edges. We use the laser rangefinder to compensate the depth value on object edges, which achieves depth error$<1\text{\textperthousand}~$ in $3~m$. We horizontally align the laser rangefinder with RS-D435i in the setup. In data acquisition, we first capture aligned RGB and depth maps through the Intel SDK~\cite{RealSense-SDK}. Then use the laser rangefinder to calibrate the depth on the object edges.

\textbf{Pixel-level alignment of images.} To align multi-camera images at the pixel level, we need to address (a). occlusion caused by different camera perspectives, and (b). pixel displacements caused by varying lens distortions across cameras.

\textit{(1) Occlusion-avoid multi-camera alignment.} The data collection approaches~\cite{benavides2022phonedepth,gao2021joint} for existing datasets fix multiple cameras horizontally. However, this setup introduces region occlusions in the images captured by these cameras, resulting in the absence of ground truth. To address this issue, we place each camera on different sliders and adjustable brackets and adjust their position to align the center of view for multiple cameras. We use a crosshair calibration board to validate the alignment. It is deemed successful if the variation in crosshair center positions captured by all cameras is under two pixels. 



 \textit{(2) RS-ToF-Stereo image registration.} 
Due to varying perspective distortions of the camera lens, mere cropping or translation of images still brings pixel displacement. It can severely compromise the precision of model training. To address this issue, we firstly acquire the parameters\footnote{The parameters refer to the camera intrinsic parameters, rotation matrix, and translation vectors.} of all cameras. In particular, we use the hollow circle calibration board (Figure~\ref{2.5D pattern}) to get the parameters of mobile ToF. Since the limited power of mobile ToF leads to pixel gaps on the circle edges of the board, we use a time filter to merge a sequence of consecutively captured frames (60 frames) into a cohesive composite frame. It effectively fills in the missing pixel along the circle edges, resulting in an accurate assessment of parameters. After that, we convert the pixel coordinates of RS-D435i and Mobile ToF images to match the pixel coordinates of the RGB camera based on these parameters. Then, we align the pixel-level images within the same coordinate system.

\section{Implementation}\label{implementation}
The core code of our system, including system deployment and dataset collection, consists of a total of 6050 lines, excluding model training and testing code.

\subsection{System Preparation and Deployment}
\textbf{Camera calibration.}
For the calibration of the stereo RGB and RealSense cameras, we use the OpenCV~\cite{culjak2012brief} calibration tool based on Zhang's method~\cite{zhang1999flexible}, while the calibration of the ToF camera is performed using the circular centroid calibration method~\cite{rudakova2014camera}. Each camera's calibration involved capturing over 20 images and filtering out image pairs with reprojection errors exceeding 0.1 pixels.
We utilize the \textit{cvStereoRectify} function from OpenCV to perform epipolar rectification on the stereo camera setup. 


\textbf{Model conversion.}
We convert the trained models to the MNN framework~\cite{MNN} and export them as separate models in FP32 and FP16 data types. For deployment on the NPU, we utilized the OMG tool from HUAWEI HiAI Engine~\cite{HiAI} to convert the MNN models into IR models. 

\textbf{System deployment.} We  implement \sysname on commodity Android mobile devices in Java~\cite{arnold2005java} and C++~\cite{stroustrup2013c++}. Specifically, we use the \textit{camera2 API} to individually access the dual RGB cameras and ToF camera, respectively.  In the Android Studio project, we include the pre-compiled \textit{libMNN.so} library file, \textit{huawei-hiai-vision.aar} and \textit{huawei-hiai-pdk.aar} for HUAWEI HiAI Engine to use mobile NPU. 

\subsection{Parallel Execution}\label{parallel}
To improve the runtime efficiency of \sysname, we identify the opportunities for parallel execution of the modules: 
\begin{itemize}
\item The stereo matching and SDEI modules require sequential operation, with the SDEI module depending on the input generated by stereo matching. Moreover, deploying these two modules on the same processor eliminates IO latency overhead resulting from data copying.
\item The TDEI module and stereo matching can run in parallel on separate computing units simultaneously, as they operate independently.
\item The Fusion module as the final module of the model, needs to run after the preceding modules have completed their execution.   
\end{itemize}

We adopt an \textit{\textbf{adaptive parallel execution}} approach to schedule modules to run in parallel on suitable processors and consider the varying computing capabilities of different mobile devices.  When the system is initially deployed on a mobile device, we search for available heterogeneous processors, currently supporting CPU, GPU, and NPU. Each module is assigned to a specific processor according to the minimized execution time. Specifically, during our testing on Huawei P40Pro and Mate30Pro smartphones, the TDEI module and Fusion module run on the NPU\footnote{The NPU supports FP16 data type models.}, while the stereo matching and SDEI modules are deployed on the GPU. On the Samsung S20+, the TDEI modules run on the CPU, while the remaining modules run on the GPU. 
\section{Evaluation}\label{sec_evaluation}
\begin{figure*}[t!]
\centering
\includegraphics[width=1\linewidth]{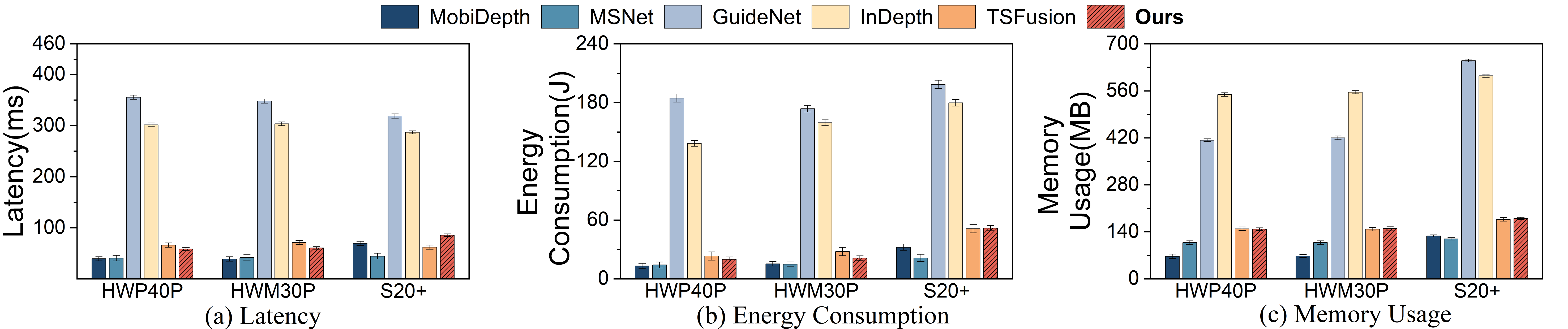} 
\caption{Performance comparison of \sysname and the baselines on different mobile devices.}
\label{comparison_MobiFuse}
\end{figure*} 

\subsection{Experimental Setup}\label{setup}
\textbf{Hardware platform.} We implement \sysname on three types of smartphones, e.g., Huawei (HW) P40P, Huawei Mate30P and Samsung S20+, each utilizing diverse SoCs and incorporating unique variants of ToF sensor, as shown in Table~\ref{Hardware}. 

\begin{table}[h!]
\centering
\caption{Hardware configurations of the mobile devices used in the experiments.}
\begin{threeparttable}
		\setlength{\tabcolsep}{1.5mm}{
\begin{tabular}{cccc}
	\toprule[1.0pt]
				\textbf{Device}\tnote{$*$} & \textbf{SoC} & \textbf{ToF} &\textbf{Support NPU}  \\  \midrule[0.5pt]
				HWM30P & Kirin 990 & Sony IMX 316 &Yes  \\
				HWP40P & Kirin 990 & Sony IMX 516  & Yes\\
				Samsung S20+ & Snapdragon865 & Sony IMX 516& No  \\
				\bottomrule[1.0pt]
\end{tabular}}
\begin{tablenotes}
		\footnotesize
		\item[$*$]HWM30P and HWP40P are the abbreviations of Huawei Mate30Pro and Huawei P40Pro, respectively.
	\end{tablenotes}
\end{threeparttable}
\label{Hardware}
\end{table}

\textbf{Baselines.} We select two mobile stereo-matching systems, e.g., MobiDepth~\cite{zhang2022mobidepth} and MSNet~\cite{shamsafar2022mobilestereonet}, two SoTA depth completion systems, e.g., GuideNet~\cite{tang2020learning} and Indepth~\cite{zhang2022indepth}, and the ToF-Stereo fusion system, e.g., TSFusion~\cite{agresti2017deep}\footnote{Due to unavailability of their source code, we re-implemented the system followed by their core idea.}, as primary baselines for comparison with \sysname. All the baseline models are obtained from their open source and retrained on our training dataset for a fair comparison. Especially, the MSNet is lightweight based on its original framework.  For the accuracy test, we also compare a \textit{Simple Fusion} 
(\textit{\textbf{SF}}) method that only filled the missing regions of the  ToF depth map with depth from stereo-matching, without comparing the depth errors of ToF and stereo-matching. 

\textbf{Datasets.} We use \textit{RealToF} for model training and testing. Among these, we randomly chose 100 images for the test set. 
To further verify the system's generalization, we performed testing on the REAL3 dataset~\cite{agresti2019stereo} without training. We manually aligned the dataset and standardized its resolution to match the testing parameters of our system.



\textbf{Metrics.} We use the following metrics to evaluate the accuracy of \sysname~\cite{zhang2018deepdepth}, including the mean absolute error(MAE) and the root mean squared error(RMSE):
\begin{equation}
\begin{aligned}
    MAE&=\frac{1}{n}\sum_{i=0}^{n}\left \| D_{p}(i)-D_{gt}(i) \right \| \\
    RMSE&=\sqrt{\frac{1}{n}\sum_{i=0}^{n}\left \| D_{p}(i)-D_{gt}(i) \right \|^{2}}
\end{aligned}
\end{equation}
We also use the percentages of pixels with predicted depths falling within an interval $\delta$ to evaluate the depth error, which is defined as following $max\left(\frac{D_{p}}{D_{gt}},\frac{D_{gt}}{D_{p}}\right )<\delta \nonumber$, where $\delta$ is $1.05, 1.10, 1.25$, respectively. These metrics were derived with the set
of samples and the evaluation script in~\cite{huang2019indoor}. 

To evaluate the accuracy of 3D reconstruction, we use Chamfer distance (\textbf{\textit{CD)}}~\cite{wu2021density} and Hausdorff distance (\textbf{\textit{HD}})~\cite{aspert2002mesh}, to measure the similarity between the prediction and ground truth. During the precision evaluation of 3D semantic segmentation, we employ two common metrics: mean Intersection over Union (\textbf{\textit{mIoU}})~\cite{Yu_2016} and Pixel accuracy (\textbf{\textit{Pixel Acc.}}).


\subsection{Overall Performance}
\label{Perf_of_DRFDeep}
In this section, we first conduct accuracy comparisons between \sysname and all the baselines on the dataset. Then, we separately compare our system with depth fusion systems and stereo-matching algorithms in terms of latency, power consumption, energy consumption, and memory usage.

\subsubsection{\bfseries MobiFuse accuracy versus all baselines}\label{accuracy}
\
\newline
As shown in Table~\ref{RealToF-Testing},  our system surpasses the baselines in all evaluated metrics on the \textit{RealToF}. 
\begin{table*}[t]
\centering
	\renewcommand\arraystretch{1.2} 
	\caption{The depth error(unit:$mm$) of baselines and our MobiFuse for the \textit{RealToF} and \textbf{REAL3} dataset. }
	\label{RealToF-Testing}
	\begin{threeparttable}
\setlength{\tabcolsep}{3.5mm}{
\begin{tabular}{l|ccccc||ccccc}
\toprule[1.0pt]
\multirow{2}{*}{Method} & \multicolumn{5}{c||}{RealToF}                                 & \multicolumn{5}{c}{REAL3~\cite{agresti2019stereo}}         \\ \cline{2-11} 
                        & MAE$\downarrow$ & \multicolumn{1}{c|}{RMSE$\downarrow$} & 1.05$\uparrow$ & 1.10$\uparrow$ & 1.25$\uparrow$ & MAE$\downarrow$ & \multicolumn{1}{c|}{RMSE$\downarrow$} & 1.05$\uparrow$ & 1.10$\uparrow$ & 1.25 $\uparrow$\\ \hline
MobiDepth~\cite{zhang2022mobidepth}   & 93.88   & \multicolumn{1}{c|}{130.93}  &  0.655    &   0.804   &   0.880      & 99.17  & \multicolumn{1}{c|}{146.68}   & 0.631     & 0.774     & 0.812     \\
MSNet~\cite{shamsafar2022mobilestereonet}                  & 57.32   & \multicolumn{1}{c|}{90.49}    & 0.745     & 0.898     & 0.942      & 90.45   & \multicolumn{1}{c|}{112.41}    &  0.657    & 0.813     &  0.883    \\
GuideNet~\cite{tang2020learning}                & 146.42  & \multicolumn{1}{c|}{198.70}   &  0.481    &  0.649    & 0.814      & 529.23  & \multicolumn{1}{c|}{926.73}   &   0.298   &   0.363   &  0.532    \\
InDepth~\cite{zhang2022indepth}                 & 54.67   & \multicolumn{1}{c|}{87.34}   & 0.775     &  0.902    & 0.946      & 248.45  & \multicolumn{1}{c|}{398.34}   & 0.417     &   0.499   &    0.578  \\
TSFusion~\cite{agresti2017deep}                & 80.49   & \multicolumn{1}{c|}{128.34}   &  0.694      & 0.835       & 0.918     & 95.77        & \multicolumn{1}{c|}{157.83}         &  0.801    &   0.849   & 0.903     \\
SF                      & 43.63   & \multicolumn{1}{c|}{74.80}    &0.839 &0.921 &0.954      & 41.65   & \multicolumn{1}{c|}{69.48}    &  0.828    &  0.903    &  0.931    \\ \hline
\textbf{MobiFuse}   & \textbf{32.62}   & \multicolumn{1}{c|}{\textbf{56.45}}     &  \textbf{0.902} &  \textbf{0.970}    & \textbf{0.982}         & \textbf{38.57}   & \multicolumn{1}{c|}{\textbf{62.32}}    & \textbf{0.869}     &  \textbf{0.939}    & \textbf{0.962}    \\
\bottomrule[1.0pt]
\end{tabular}}
\end{threeparttable}
\end{table*}
For instance, compared to MobiDepth and MSNet, \sysname achieves a remarkable reduction of 65.3\% and 43.1\% in MAE, respectively. Likewise, our system demonstrates a significant decrease of 77.7\%, 40.3\%, and 59.5\% compared to GuideNet, InDepth and TSFusion in MAE, respectively.  For $\delta$ percentage metric,  \sysname also achieves the highest value compared to all the other baselines. Moreover, compared with \textit{SF} method, \sysname can further reduce the depth error with 25.2\% in the MAE and 24.5\% in the RMSE. This is because the \textit{SF} method cannot compare the depth error between stereo-matching and ToF, leading to the suboptimal choice. It further highlights the significance of \sysname to compare the depth errors between ToF and stereo matching at each pixel.   



\subsubsection{\bfseries Generalization of the system}
\
\newline
Our system's generalization performance is evaluated on the REAL3 dataset. As indicated in Table~\ref{RealToF-Testing}, our system achieves impressive results even without prior training with an MAE of 38.57$mm$. This outperforms other methods, including GuidNet and InDepth, which rely on RGB features for depth completion and exhibit MAE values of 529.23$mm$ and 248.45$mm$, respectively. By directly learning the objective physical relationship between environmental factors and depth errors from ToF and binocular vision, our system significantly improves generalization performance due to the consistent nature of these physical relationships across most scenarios. Furthermore, our system exhibits higher accuracy compared to methods, such as TSFusion and \textit{SF}. Notably, \textit{SF} performs slightly better on the REAL3 dataset than on \textit{RealToF} and is close to \sysname. This can be attributed to the superior performance of ToF in the REAL3 dataset, which exhibits fewer depth deficiencies. As a result, the \textit{SF} predominantly relies on ToF for depth estimation. However, in rare cases where stereo matching still outperforms ToF in specific regions,  \sysname~can accurately compare the depth errors between the two methods, selecting the most reliable depth values and enhancing the overall depth perception precision.

\subsubsection{\bfseries Performance comparison with baselines}
\
\newline
\sysname outperforms both GuideNet and InDepth, with its lightweight model and efficient parallel deployment strategy, offering superior performance in terms of latency, energy consumption, and memory usage. Specifically, \sysname has significantly reduced latency by 84.4\% and 81.6\% on HWP40P, as shown in Figure~\ref{comparison_MobiFuse}(a). Compared to TSFusion, our system exhibits lower latency on Huawei smartphones equipped with NPU. On Samsung S20+, \sysname introduces a slightly higher latency of 7 ms compared to TSFusion. However, considering the significant improvement in accuracy achieved by our system, this slight increase in latency is entirely acceptable. 

For testing the energy consumption, we give a set of computation tasks, i.e., 100 pairs of images. On HWM30P, \sysname demonstrates energy savings of 87.7\% and 85.3\% compared to GuideNet and InDepth, respectively,  and it slightly outperforms TSFusion, as shown in Figure~\ref{comparison_MobiFuse}(b). We measure the memory usage of the two baselines and \sysname using the Monitors tool in Android Studio. As shown in Figure~\ref{comparison_MobiFuse}(c), \sysname shows superior efficiency in memory utilization on the Samsung S20+. 

Compared to the two stereo matching methods on mobile devices, MobiDepth and MSNet, \sysname incurs slight overheads since the stereo matching is incorporated as a component to provide disparity and cost volume inputs for the SDEI module.  However, given that \sysname improves the accuracy remarkably over the stereo matching methods and the latency is sufficient to meet the responsiveness requirements of the majority of depth-based applications on mobile devices, i.e., a low latency of only 55.3$ms$ (18FPS) on HWP40P, we believe that the added overhead is acceptable. 



\subsection{Micro Benchmarks}
\subsubsection{\bfseries Performance of TDEI and SDEI module}
\begin{figure}[h!]
    \centering
    \includegraphics[width=1\linewidth]{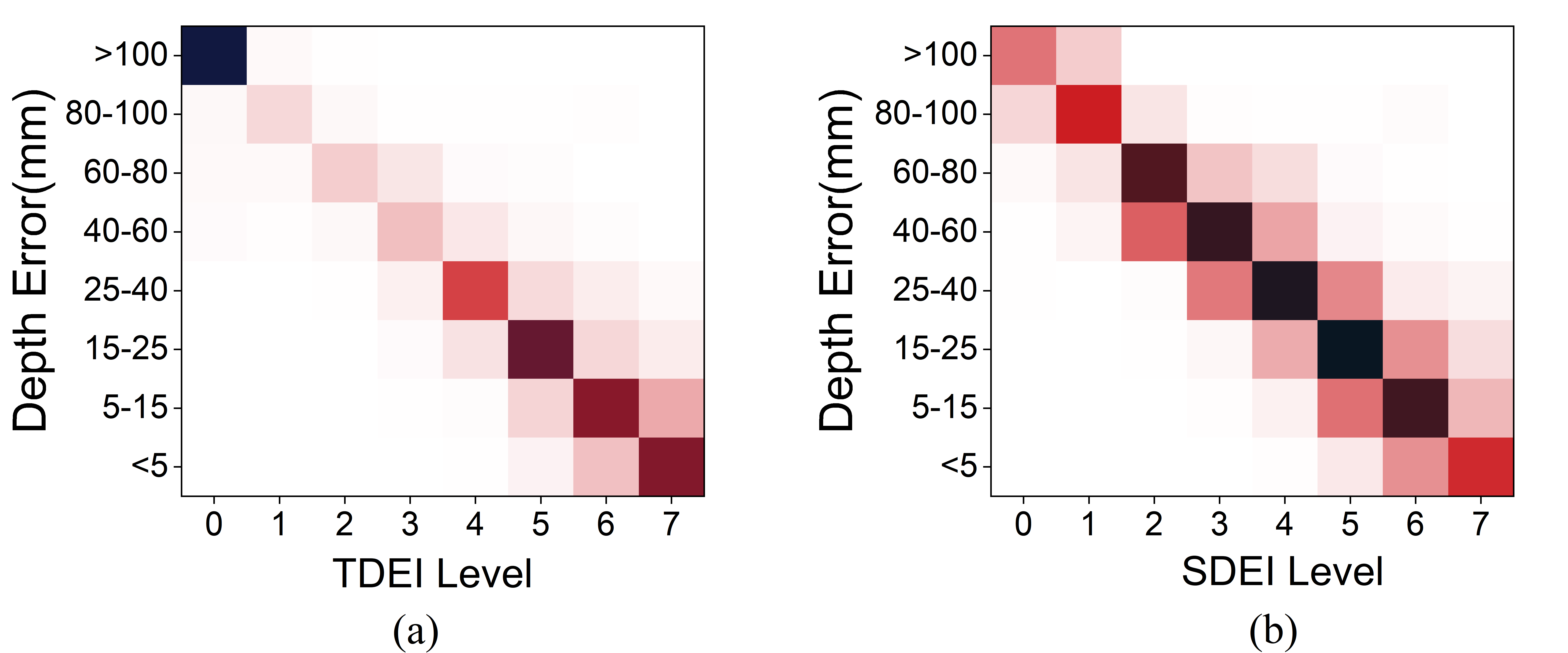}

    \caption{Correlation analysis between depth errors and DEI metric proposed by our approach. }
  \label{DEI-error-test}
\end{figure}

\
\newline
We separately test the TDEI and SDEI modules to gauge their precision in recognizing DEI levels, as shown in Figure~\ref{DEI-error-test}, which showcases the correlation distribution between TDEI, SDEI, and depth error. The TDEI module demonstrates 80.4\% in Top-1 accuracy, while the SDEI module achieves 70.3\%. In Top-2 accuracy, the TDEI module excels at 88.4\% and the SDEI reaches 81.5\%. 
Figure~\ref{DEI-error-test} clearly shows a linear correlation between both modules' DEI and the corresponding depth error. This compelling relationship proves the accuracy of our DEI in effectively capturing the magnitude of depth error.

\subsubsection{\bfseries Compatibility on different mobile devices}
\
\newline
\sysname is trained on the \textit{RealToF} collected by HWP40P. To assess the system's compatibility with other mobile devices, we captured 60 samples using HWM30P and Samsung S20+, respectively, for accuracy testing. As shown in Table~\ref{Samsung_test}, \sysname consistently outperforms all baselines on these samples, indicating satisfactory device compatibility. Due to limited space, only MAE and RMSE metrics are shown.

\begin{table}[t]
	\renewcommand\arraystretch{1.2} 
	\caption{Comparison of \sysname on different devices.}
 	\label{Samsung_test}
	\centering
 \begin{threeparttable}
\setlength{\tabcolsep}{1.8mm}{
\begin{tabular}{l|cc||cc}
\toprule[1.0pt]
\multirow{2}{*}{Method} & \multicolumn{2}{c||}{Samsung S20+}                     & \multicolumn{2}{c}{Huawei Mate30 Pro}                     \\ \cline{2-5} 
& MAE($mm$)$\downarrow$ & RMSE($mm$)$\downarrow$ & MAE$\downarrow$ & RMSE$\downarrow$  \\ \hline
MobiDepth               & 128.51   & 195.16        &  108.98    & 141.19      \\
MSNet                   & 80.13    & 100.95        &  45.31    & 63.33        \\
GuideNet                & 195.09   & 268.43      &  169.94   & 223.42       \\
InDepth                 & 115.44   & 195.41       &  63.28    & 80.72         \\ 
TSFusion              &     161.95       &   329.61          &    90.77        &        167.83   \\
\textit{SF}          & 65.63    & 113.89    &  51.56    &  68.88          \\ \hline
\textbf{Ours}           & \textbf{59.15}    & \textbf{93.88}   &   \textbf{41.32}   &  \textbf{60.55}   \\
\bottomrule[1.0pt]
\end{tabular}}
\end{threeparttable}
\end{table}

\subsubsection{\bfseries Impact of DEI category quantity.}
\
\newline
A proper classification of DEI labels can enhance the accuracy of \sysname. To achieve this, we test the final accuracy of the system under DEI classifications with 4, 8, and 16 respectively. DEI-8 follows the classification guidelines provided in Table~\ref{DEI Level}, while DEI-4 and DEI-16 involve merging and decomposing categories based on the DEI-8 classification. As shown in Table~\ref{DEI_Test}, the system based on DEI-8 provides a depth error reduction of 14.5\% and 13\% compared to DEI-4 and DEI-16, respectively. Therefore, training the TDEI and SDEI in our system using an 8-level DEI approach is reasonable.
\label{label_test}
\begin{table}[h]
\setlength{\abovecaptionskip}{0pt}
	\setlength{\belowcaptionskip}{1pt}
	\renewcommand\arraystretch{1.2} 
	\caption{Comparison of accuracy for different numbers of DEI label categories.}
	\label{DEI_Test}
	\centering
 \begin{threeparttable}
\setlength{\tabcolsep}{2mm}{
\begin{tabular}{l|cc|ccccc}
\toprule[1.0pt]
\textbf{Method}      & \textbf{MAE($mm$)}$\downarrow$ & \textbf{RMSE($mm$)}$\downarrow$ & \textbf{1.05}$\uparrow$ & \textbf{1.10}$\uparrow$ & \textbf{1.25}$\uparrow$  \\ \hline
DEI-4   & 38.13   &  69.10  & 0.844    & 0.940     & 0.968     \\
\textbf{DEI-8}    & \textbf{32.62}   & \textbf{56.45} &  \textbf{0.902} &  \textbf{0.970}    & \textbf{0.982}    \\
DEI-16  & 37.48   &  68.24    & 0.848    & 0.943     & 0.969         \\ 
\bottomrule[1.0pt]
\end{tabular}}
\end{threeparttable}
\end{table}

\subsection{Case Study}
\sysname enables various depth-based applications on mobile devices. We take Depth-based 3D reconstruction and RGB-D-based 3D segmentation as case studies to show the effectiveness of \sysname. 

\subsubsection{\bfseries Case1: Depth-based 3D reconstruction} 
\
\newline
We utilize 50 depth maps obtained from \sysname and the baselines as inputs to InfiniTAM~\cite{InfiniTAM_arXiv_2017} for 3D reconstruction, with the depth obtained by RealSense serving as the ground truth. Table~\ref{3D reconstruction}  reveals that the 3D reconstruction results based on \sysname's depth outperform those of the baselines, with the $CD$ metric of 0.03 and the $HD$  metric of 0.21. 
\begin{table}[h!]
\centering
\caption{Comparison of 3D reconstruction accuracy for different methods of depth input.}
\begin{threeparttable}
\setlength{\tabcolsep}{0.8mm}{
\begin{tabular}{c|ccccccc}
	\toprule[1.0pt]
Methods            & MobiDepth     & MSNet    & GuideNet   & InDepth  & TSFusion &\textit{SF}  & \textbf{Ours} \\ \hline
$CD$$\downarrow$   &  0.21         & 0.06     & 0.17       & 0.06   & 0.05  & 0.04   &  \textbf{0.03} \\
$HD$$\downarrow$   &  0.88         & 0.30     & 0.39       & 0.33   & 0.31  & 0.24  &   \textbf{0.21}   \\
        \bottomrule[1.0pt]
\end{tabular}}
\end{threeparttable}
\label{3D reconstruction}
\end{table}

\begin{figure*}[t]
	\centering
	\includegraphics[width=0.85\linewidth]{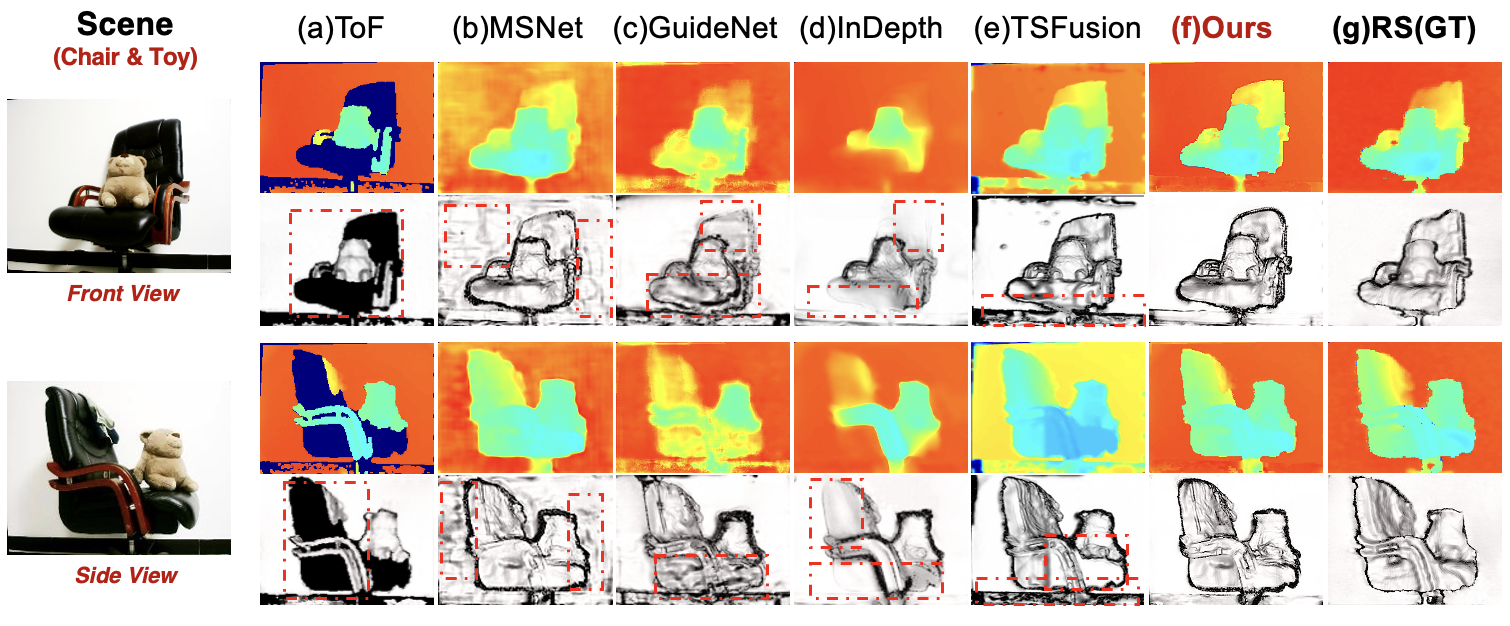}
	\caption{The first row shows depth maps obtained from various depth measurement methods and \sysname.  Using InfiniTAM~\cite{InfiniTAM_arXiv_2017}, the second row shows the 3D reconstructions from depths obtained by different methods as input.} 
	\label{3D re}
\end{figure*}

In the qualitative results in Figure~\ref{3D re},  we highlight, using red dashed boxes, areas where inaccurate input depth maps lead to suboptimal 3D reconstruction results, from left: (a) mobile ToF, (b) MSNet, (c) GuideNet, (d) InDepth, (e) TSFusion, (f) our \sysname, (g) 
the ground truth (GT) captured by RealSense D435i. As depicted in the figure, the absence of depth information directly hampers object reconstruction in 3D. Moreover, inaccuracies in input depth result in issues like pixel misalignment and overlap on the object surfaces during 3D reconstruction. The illustration also confirms that the more reliable depth data provided by \sysname enhances the effectiveness of object 3D reconstruction.   


\subsubsection{\bfseries Case2: RGB-D-based 3D segmentation} 
\
\newline

\begin{figure*}[t]
	\centering
	\includegraphics[width=0.8\linewidth]{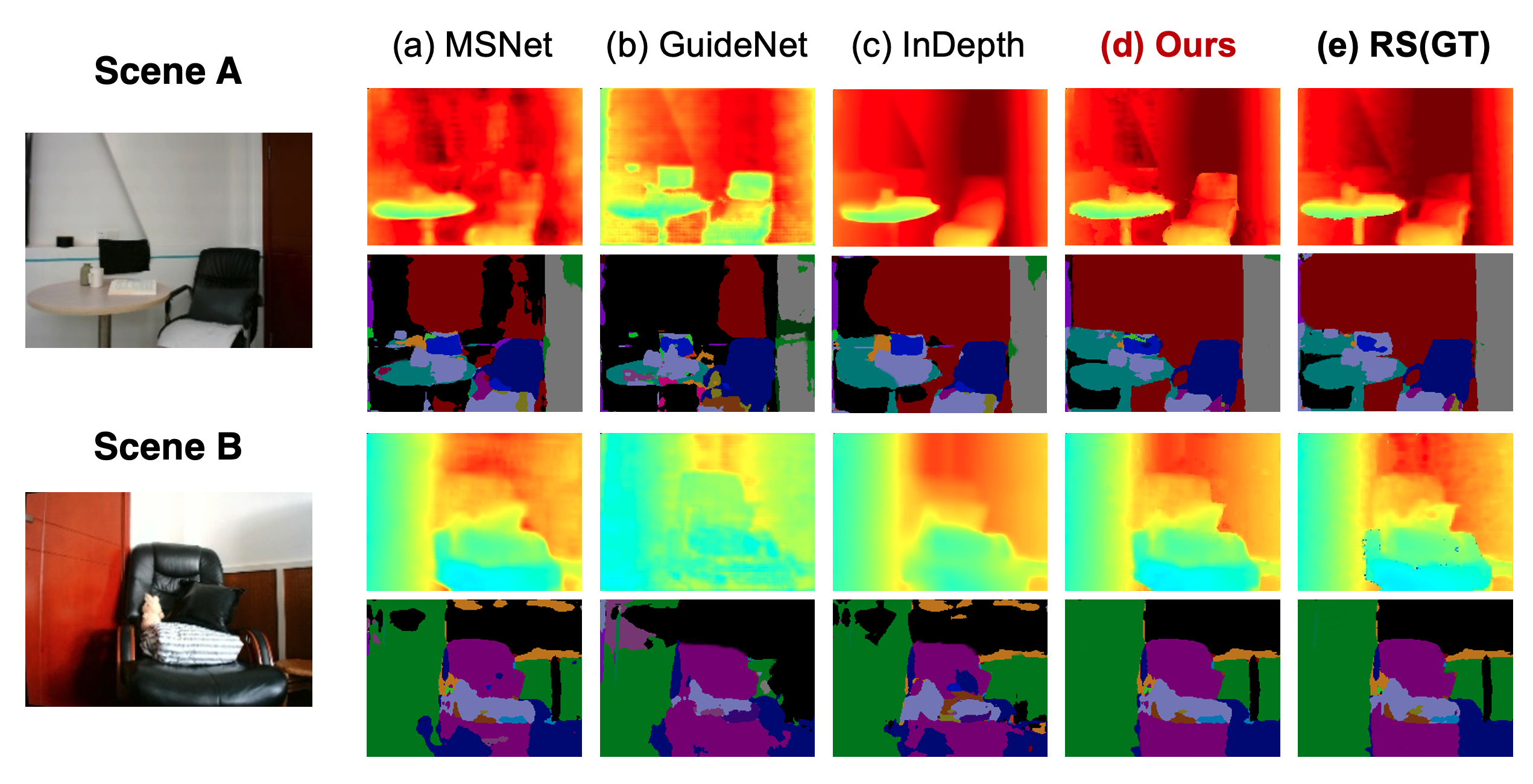}
	\caption{The top row displays depth maps. The bottom row presents 3D segmentation results generated by the CMX\cite{zhang2023cmx} using depth maps acquired through different methods as input. } 
	\label{3D seg}
\end{figure*}

To further validate the significance of high-precision depth data for downstream 3D applications, we input 50 depth maps obtained from MSNet, GuideNet, InDepth, and MobiFuse in different scenarios into the state-of-the-art RGBD-based 3D segmentation model CMX~\cite{zhang2023cmx} for accuracy testing, as shown in Table~\ref{3D segmentation}. The experimental results indicate that the 3D segmentation accuracy using MobiFuse depth maps as input outperforms other comparative methods. The \textbf{\textit{mIoU}} increased by a minimum of 21.3\%, and the \textbf{\textit{Pixel Acc.}} also increased by a minimum of 8.3\%. This underscores the crucial importance of high-precision depth maps for existing 3D applications. 
\begin{table}[t]
\centering
\caption{Comparison of 3D segmentation accuracy for different methods of depth input.}
\begin{threeparttable}
\setlength{\tabcolsep}{3.4mm}{
\begin{tabular}{c|cccc}
	\toprule[1.0pt]
Methods        & MSNet & GuideNet & InDepth & \textbf{Ours} \\ \hline
mIoU(\%) $\uparrow$      & 51.46 & 20.84    & 50.79   & \textbf{62.43}    \\
Pixel Acc.(\%)$\uparrow$ & 85.82 & 60.92    & 85.24   & \textbf{92.96}    \\
        \bottomrule[1.0pt]
\end{tabular}}
\end{threeparttable}
\label{3D segmentation}
\end{table}

Visualizations of the 3D segmentation effects are depicted in Figure~\ref{3D seg}, showing that \sysname depth precision is the highest, resulting in optimal 3D segmentation outcomes.


\section{Related Work}\label{sec_related_work}

\textbf{ToF-based depth fusion methods.} Some fusion approaches use CNN models to combine RGB images and  ToF depth maps to enhance depth estimation accuracy~\cite{zhang2022indepth,tang2020learning,zhang2018deepdepth}. Yet they necessitate dependable input images, exhibit poor generalization, and impose notable latency on mobile devices. Other methods~\cite{agresti2017deep, poggi2019confidence} calculate depth fusion weights based on ToF and stereo confidence. However, 
these methods use received signal amplitude strength to predict the ToF confidence, which can only reflect the probability distribution of ToF depth errors, without directly characterizing the actual magnitude of the depth errors. Moreover, they fail to fully exploit the correlation and complementarity between the information of ToF and stereo-matching in depth fusion, leading to imprecise depth fusion accuracy.

In \sysname, we use physical principles from environmental factors to propose Depth Error Indication (DEI) modal information, precisely indicating depth errors in ToF and stereo-matching. We employ the progressive fusion approach with backward connection to combine geometric properties from ToF and stereo depth maps with depth error features from the DEI modality to craft accurate fused depth maps.

\section{Discussion}
\label{discussion}

\noindent \textbf{Enhancing edge pixel depth perception in \sysname.}  
Both ToF and stereo matching are limited to accurately measuring the depth of some intricate contours, resulting in suboptimal depth results for edge pixels in \sysname. While \sysname utilizes pixel depth gradients to identify edge pixels, it struggles to distinguish whether these pixels belong to the foreground or background. In our ongoing work, we aim to improve this by incorporating RGB images into \sysname and refining the depth of ambiguous edge pixels through object segmentation and depth interpolation.

\section{Conclusion}\label{sec_conclusion}
In this paper, we propose the ToF-stereo depth fusion system, \sysname, to complementarily leverage the on-device dual cameras and ToF. Additionally, we create the \textit{RealToF} dataset to fill the gap in current datasets that focus on ToF-Stereo fusion in real-life scenarios. Experiments show that \sysname significantly outperforms existing depth measurement methods and can work on mobile devices efficiently.

\bibliographystyle{IEEEtran}
\bibliography{main}

\end{document}